\def\year{2021}\relax
\documentclass[letterpaper]{article}
\usepackage{aaai21} 
\usepackage{times} 
\usepackage{helvet} 
\usepackage{courier}
\usepackage[hyphens]{url}
\usepackage{graphicx}
\usepackage{natbib}
\usepackage{caption}
\usepackage{amsmath,amssymb}
\usepackage{xcolor}
\usepackage{array} 
\usepackage{mathtools}

\definecolor{ylgnbu8c7}{HTML}{2432FF}
\definecolor{ylgnbu9c9}{HTML}{081D58}
\definecolor{blues7c7}{HTML}{084594}
\definecolor{greens7c5}{HTML}{41AB5D}
\definecolor{paired10c4}{HTML}{33A02C}
\definecolor{paired10c6}{HTML}{E31A1C}
\definecolor{ylorrd3c2}{HTML}{FEB24C}
\definecolor{5thcolor}{HTML}{FF0080}

\usepackage[switch]{lineno}
\usepackage{accents} 
\newcommand\bb[1]{\underaccent{\bar}{#1}}

\newcommand{\ub}[2]{\bar{#1}_{#2}} 
\newcommand{\lb}[2]{\bb{#1}_{#2}}
\newcommand{\ab}[2]{\tilde{#1}_{#2}}

\newcommand{\beginsupplement}{%
        \setcounter{table}{0}
        \renewcommand{\thetable}{S\arabic{table}}%
        \setcounter{figure}{0}
        \renewcommand{\thefigure}{S\arabic{figure}}%
     }

\newtheorem{definition}{Definition}

\title{Understanding Interpretability by generalized distillation in  Supervised Classification}
\author {
        Adit Agarwal,\textsuperscript{\rm 1}
        Dr K.K.Shukla, \textsuperscript{\rm 1}
        Arjan Kuijper,  \textsuperscript{\rm 2}
        Anirban Mukhopadhyay \textsuperscript{\rm 3} \\
}
\affiliations {
    Affiliations
    \textsuperscript{\rm 1} IIT (BHU) Varanasi, India \\
    \textsuperscript{\rm 2} IGD Fraunhofer, TU Darmstadt, Germany \\
    \textsuperscript{\rm 3} GRIS, TU Darmstadt, Germany \\
}
\begin{document}
\maketitle
\begin{abstract}
The ability to interpret decisions taken by Machine Learning (ML) models is fundamental to encourage trust and reliability in different practical applications. Recent interpretation strategies focus on human understanding of the underlying decision mechanisms of the complex ML models. However, these strategies are restricted by the subjective biases of humans. To dissociate from such human biases, we propose an interpretation-by-distillation formulation that is defined relative to other ML models. We generalize the distillation technique for quantifying interpretability, using an information-theoretic perspective, removing the role of ground-truth from the definition of interpretability. Our work defines the entropy of supervised classification models, providing bounds on the entropy of Piece-Wise Linear Neural Networks (PWLNs), along with the first theoretical bounds on the interpretability of PWLNs. We evaluate our proposed framework on the MNIST, Fashion-MNIST and Stanford40 datasets and demonstrate the applicability of the proposed theoretical framework in different supervised classification scenarios. 
\end{abstract}
\section{Introduction}
Trust and reliability form the core principles in various safety-critical applications such as {medicine} and autonomous driving. With the growing demand for more complex data-driven computational models for such applications, model interpretation has become the focus of major research in the past few years \cite{46160}.

The deployment of deep learning models in critical domains, such as law and medicine, requires unbiased computational estimates of interpretability. This would allow policy-makers in these domains to evaluate and select computationally more interpretable models for different tasks, free from subjective human biases. 

The motivations and difficulties associated with interpretability are explained by \cite{Lipton:2018:MMI:3236386.3241340}.
The lack of a single notion of interpretability leaves the applications of deep learning models vulnerable, thus amplifying the need for a robust quantification of interpretability.
This lack of a proper measurable bias-free definition for interpretability poses a serious problem when deep learning models fail silently, leaving end-users with no clues on the possible correction mechanisms, which can be fatal in safety-critical scenarios.

Inspired by the work of \cite{DBLP:journals/corr/abs-1806-10080}, we define interpretability from an information-theoretic perspective and decouple human understanding from the current notion of interpretability.
We define the process of interpretation as a communication mechanism between two models and define interpretability relative to their model entropies. Since the abstraction levels defined in \cite{DBLP:journals/corr/abs-1806-10080} do not properly define the entropies of Machine Learning (ML) models, our work proposes a novel way of defining these entropies. 
Further, we derive tight lower bounds on the entropy of supervised classification models using graph theory.

Our interpretation-by-distillation framework for quantifying interpretability (refer Fig. \ref{fig:interpretation}) generalizes the common distillation technique \cite{44873}; providing researchers, corporations and policy-makers, better, unbiased interpretability estimates.
Further, we provide the first theoretical guarantees on the interpretability of black-box Piece-Wise Linear Networks (PWLNs), when interpreted by another PWLN. The major contributions of our proposed theoretical framework are:
\begin{itemize}
    \item We remove the accuracy-interpretability trade-off, present in most previous works. We propose that the interpretability of any ML model depends only on the decision structure of the model, independent of the ground-truth. 
    \item Local surrogates such as SHAP \cite{NIPS2017_7062} and LIME \cite{DBLP:journals/corr/RibeiroSG16} provide a localized view of robustness of classifiers only around individual data points. Our work, on the other hand, proposes a global metric for defining the interpretability of a model in relation to another model.
    \item Our empirical interpretation-by-distillation mechanism represents a computational approach for interpretability. We generalize standard distillation which considers the generic human understanding of a lower complexity model being used to interpret a higher complexity model, by considering the entire spectrum of computational complexity of learning models. 
\end{itemize}

\section{Related Work}
The necessity for humans to have confidence on the predictions of deep learning models \cite{EU} has led to the development of various explanation mechanisms \cite{DBLP:journals/corr/abs-1802-01933}. These mainly explore two directions - Model-based and Post-hoc methods.

Model-based interpretation mechanisms such as  focus on building interpretable ML models from the bottom up using basic decision mechanisms, retaining the complexity of deep neural networks, while making it easier to interpret their decisions. 
\cite{InterpretableAlzheimers}, \cite{murdoch2019interpretable}, \cite{Caruana:2015:IMH:2783258.2788613}, \cite{Abdul:2018:TTE:3173574.3174156} have explored model-based methods extensively, but they have not been able to perform at par with existing complex deep learning models.

Post-hoc interpretation mechanisms, on the other hand, have mainly focused on the visualization of deep neural networks such as CNNs \cite{DBLP:journals/corr/abs-1802-00614}.
Previous works by \cite{Kim2018InterpretabilityBF}, \cite{DBLP:journals/corr/SelvarajuDVCPB16}, \cite{DBLP:journals/corr/ShrikumarGK17}
use visual cues such as Concept Activation Vectors (CAV), Grad-CAM and Layer-wise relevance scores \cite{Bach2015OnPE} 
respectively to enable human understanding of the complex ML models. 
However, these face a major challenge due to the fragile nature of the proposed interpretations, which decreases human understanding and trust in practical systems using these cues \cite{Ghorbani2018InterpretationON}.

\begin{figure*}[t]
    \centering
    \includegraphics[width = 0.8\textwidth]{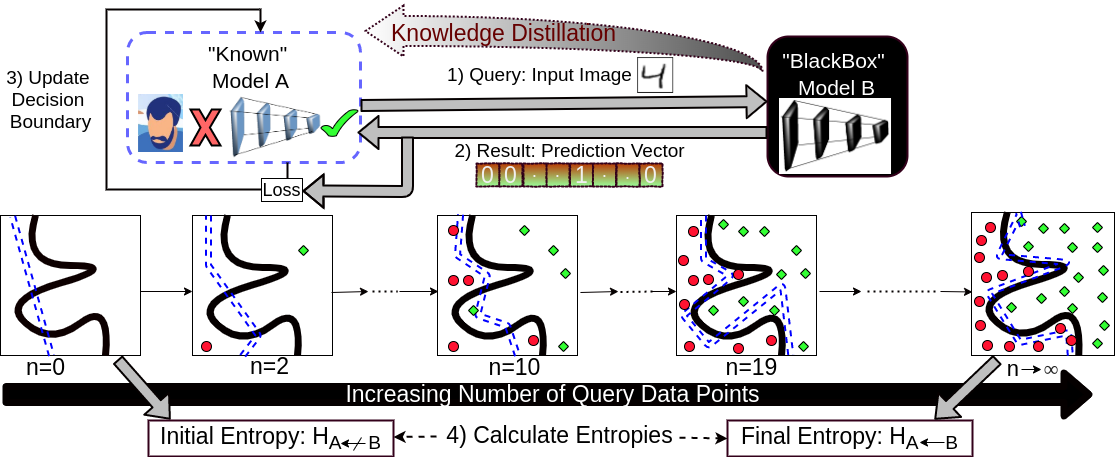}
    \caption{Interpretation as a communication mechanism between known model \textit{A} and black-box model \textit{B}, where \textit{A} performs a series of (possibly infinite) queries to \textit{B}, until it emulates \textit{B}'s decision boundary and no more information gain is possible.}
    \label{fig:interpretation}
\end{figure*}

Our interpretation-by-distillation mechanism considers interpretation by any supervised learning model, moving beyond its usual association with human understanding (see Figure \ref{fig:interpretation}).
\cite{DBLP:journals/corr/DhurandharILS17aa} present a work most similar to ours, but suffer from the accuracy-interpretability trade-off which does not arise in our work. 
We de-couple the accuracy/performance of the ML model from its interpretability.
In practical settings, interpretability can be computed by our empirical interpretation mechanism.
As compared to knowledge distillation \cite{44873}, where a small network (student) is taught by a larger pre-trained network (teacher); our interpretation-by-distillation mechanism 
covers the entire spectrum of their relative computational complexities (refer Section \ref{sect:complSpect}).\\
As a special case of interest, we consider PWLNs and derive theoretical bounds on their interpretability, in terms of their complexity defined in previous works.

\subsection{Complexity of PWLNs}
{A lot of previous research has focused on deriving the complexity of PWLNs. \cite{zaslavsky1975} first proposed the number of cells formed by an n-dimensional arrangement of hyperplanes in a d-dimensional space as $\sum_{i = 0}^{d}$ ${n}\choose{i}$. 
\cite{serra2018bounding} propose tight upper and lower bounds on the maximium number of linear regions for ReLU and maxout networks. 
\cite{NIPS2014_5422} and \cite{NotesLinearRegions} present tight lower and upper bounds respectively, on the maximum number of linear regions for ReLU networks. 
\cite{complexityAverage} provide average case complexities for ReLU networks in terms of the number of linear regions.}

While we directly use most of these results for deriving the upper and average bounds on the entropy of PWLNs, we present a tighter lower bound on the entropy by incorporating the graph coloring concept.

\section{Background}\label{sect:background}
{We consider two models \textit{A} $(\theta_{A}:\mathbf{X}\rightarrow Y_A$) and B $(\theta_{B}:\mathbf{X}\rightarrow Y_B)$ that take an $n_0$-dimensional input, say $x \in \mathbf{X} \subseteq \mathcal{R}^{n_0}$ and predict outputs $y_A$ and $y_B$ in the output space $Y_A, Y_B \subseteq \mathcal{Y}$ respectively.
{The input to both models need not be same (as is the case in Section \ref{sect:diffModelInputs})}.
These models are trained on the dataset, $\mathcal{D} \triangleq \{(x_i, y_i)\}_{i = 1}^{n}$ containing $n$ data points and $c$ output classes.

\textit{B} represents the "black-box" model being interpreted, while \textit{A} represents the "known" model used for interpreting \textit{B}, as shown in Figure \ref{fig:interpretation}. 
Model \textit{A} need not be more interpretable than \textit{B} from a human perspective. 
The interpretation process is performed using the dataset, {$\hat{\mathcal{D}} \triangleq \{(x_i, \hat{y}_i)\}_{i = 1}^{\hat{n}}$, containing $\hat{n}$ data points, where $\hat{y}_{i}$ represents the prediction of \textit{B} on the $i^{th}$ sample, $x_i$ after training.}

A \textbf{cell} formed in the input space by a model is defined as a bounded region in the input space in which all points map to the same output. For PWLNs, a \textbf{cell} indicates a bounded region where all points map to same output domain as well as have the same linear nature of the PWLN. Based on this, we derive the definition of model complexity from \cite{serra2018bounding} as in Definition \ref{complexity}.
\begin{definition}{\textbf{{Model Complexity: }}}\label{complexity}
	\textit{{The complexity of a supervised classification model is defined as the number of unique cells identified by the model in the input space.}}
\end{definition}

Throughout the paper, we denote the complexity of any model $Q$ as $C_Q$, and the upper, lower and average bounds on $C_Q$ as $\ub{C}{Q}$, $\lb{C}{Q}$ and $\ab{C}{Q}$ respectively.

The theoretical bounds on the complexity of PWLNs from previous research used for theoretical derivations are given in Supplementary Material.}

Based on Definition \ref{complexity}, we define model entropy as:
\begin{definition}{\textbf{{Model Entropy: }}}\label{entropy}
	\textit{{The entropy of a supervised classification model is defined as the number of unique ways of assigning output classes to different cells identified in the input space by the model.}}
\end{definition}
This definition of model entropy does not take into account the shape of the cells.
Throughout the paper, we denote the entropy of any model $Q$ as $H_Q$. We denote the upper, lower and average bounds on $H_Q$ as $\ub{H}{Q}$, $\lb{H}{Q}$ and $\ab{H}{Q}$ respectively.
\subsection{Relation between Complexity and Entropy}\label{sect:entropy_formmulation}
A rudimentary approximation of the relationship between the complexity and entropy of any supervised classification model, say $Q$, for a $c$-class classification problem, based on the definitions \ref{complexity} and \ref{entropy}, is given by $H_Q = (c)^{C_Q}$, representing the maximal possible number of ways of assigning the output classes.
\section{Problem Formulation}\label{sect:formulation}
As shown in Figure \ref{fig:interpretation}, we have two models \textit{A} and \textit{B}. \textit{A} belongs to the class of \textit{known} models {i.e. those models for which the decision mechanism is available to us through access to the model internals including structure and parameters}.
\textit{B} belongs to the class of \textit{black-box} models i.e. those models whose internals are completely unknown. 
{The semantic understanding of "known" is unimportant in our context.}
The interpretation mechanism is then defined as the process of communication between these models, where \textit{A} tries to emulate \textit{B}'s decision boundary. 
This process can continue indefinitely until \textit{A} has learnt all the information possible about \textit{B}'s decision mechanism.

The relative simplicity of model \textit{A} w.r.t. \textit{B} is not important in our work. Our aim is only demonstrating the explainability of \textit{B} in terms of \textit{A} and not finding the most compact model.
Based on the definitions given in \cite{DBLP:journals/corr/abs-1806-10080}, we formally define interpretability and its associated concepts as follows: 

\begin{definition}{\textbf{Complete Interpretation:}}\label{CompInterp}
\textit{Complete Interpretation is defined as the process of model updation by communication between the known model \textit{A} and the target black-box model \textit{B} through an infinite set of query data points until no further information gain about the decision boundary of model \textit{B} is possible.}
\end{definition}
To define the notion of interpretability, we first define two terms as $H_{A \not\leftarrow B}$ and $H_{A \leftarrow B}$, which represent entropies of the communication process between models \textit{A} and \textit{B}. $H_{A \not\leftarrow B}$ and $H_{A \leftarrow B}$ represent the entropy about $B$'s decision boundary before and after the process of \emph{complete interpretation} respectively. 
$H_{A \not\leftarrow B}$ and $H_{A \leftarrow B}$ can be also be obtained in terms of the model entropies $H_A$ and $H_B$ as, $H_{A \not\leftarrow B} = H_B$ and $H_{A \leftarrow B} = \max\Big((H_B - H_A), 0\Big)$.\\
{However, it is not possible to consider the entire input space $\mathcal{R}^{n_0}$ due to its massive cardinality. 
So, for practical purposes, the process of \emph{complete interpretation} (Definition \ref{CompInterp}) is relaxed into that of \emph{empirical interpretation} (Definition \ref{PartialInterp}).}
\begin{definition}{\textbf{{Empirical} Interpretation:}}\label{PartialInterp}
\textit{Empirical Interpretation is defined as the process of model updation by communication between the known model \textit{A} and the target black-box model \textit{B} using a finite subset of query data points.}
\end{definition}
{For \emph{Empirical Interpretation}, the model entropy $H_B$ cannot be determined. Empirically, the interpretability measure depends only on model \textit{A}. Hence, the model entropies $H_{A \not\leftarrow B}$ and $H_{A \leftarrow B}$ are approximated as \textit{$\mathit{H^{emp}_{A \not\leftarrow B} \approx H_A}$} and \\
\textit{$\mathit{H^{emp}_{A \leftarrow B} \approx \hat{H}_A}$}, where $H_A$ and $\hat{H}_A$ represent the entropy of model A before and after the process of \emph{Empirical Interpretation} respectively.\\
}
Based on these concepts, we define interpretability as:
\begin{definition}{\textbf{Interpretability: }}\label{interp}
	\textit{Interpretability ($I_{A \leftarrow B}$) is the ratio between information gain about target model $B$'s decision boundary through~\emph{interpretation} and the initial uncertainty about $B$'s decision boundary. 
	More formally,}
	\begin{equation} \label{interpretability}
	I_{A \leftarrow B} = \frac{H_{A \not\leftarrow B} - H_{A \leftarrow B}}{H_{A \not\leftarrow B}}, I_{A \leftarrow B} \in [0, 1] 
	\end{equation}
\end{definition}
This interpretability formulation measures the maximal information gain about the decision boundary of black-box model \textit{B}, through minimal querying by known model \textit{A}. 
Thus, an interpretability value of 0.9 means that 90\% of the decision boundary of \textit{B} can be matched by \textit{A} on a particular dataset.
This process of interpretation in Definition \ref{interp}, is modelled as the optimization given by $min(KL(p(Y_{B}|\mathbf{X}, \theta_{B})||p(Y_{B}|\mathbf{X}, \theta_{A})))$. 
This is solved by approximating $p(Y_{B}|\mathbf{X}, \theta_{A})$ from $p(Y_{A}|\mathbf{X}, \theta_{A})$, where $\theta_{B}$ is unknown and $\theta_{A}$ is known.

\subsection{Bounds on Entropy and Interpretability}
Based on the formulation in Section \ref{sect:entropy_formmulation}, we obtain the upper and average bounds on the entropy for PWLNs, in terms of the corresponding bounds on their complexities. 
However, we derive tighter lower bound estimates for these entropies.

To obtain the lower bounds, we model the entropy of PWLNs in terms of the \textit{graph coloring problem}. The linear cells identified by the PWLN in the input space can be modelled as the vertices of a graph $G$ and the output classes represent the different colors available for coloring the vertices of $G$. The entropy of the PWLN can now be modelled as the process of coloring the vertices of $G$ such that no two adjacent vertices are assigned the same color.

We model this process of calculating the entropy, as an iterative process through all the layers in the PWLN. At each iteration, the adjacent linear cells of the input space are approximated as a path graph in each dimension and the \textit{chromatic polynomial} of path graph is used to calculate the number of ways of coloring the graph. Based on the construction, we obtain a lower bound on the entropy of an L-layer PWLN, say $Q$, with $n_0$-dimensional input and $c$ output classes and $n_l$ ReLU neurons in the $l$-th layer as:
\begin{center}
    $\lb{H}{Q} = ({\displaystyle \prod_{l = 1}^{L-1}}c(c - 1)^{(\lfloor\frac{n_l}{n_0}\rfloor - 1)n_0})*(c(c-1)^{(n_L - 1)n_0})$
\end{center}
The detailed derivation of the above is given in the Supplementary Material.

\begin{table*}[!h]
\centering
{\renewcommand{\arraystretch}{1.5}
\begin{tabular}{|p{1.1cm}|p{1.3cm}p{1.3cm}p{1.3cm}|p{1.3cm}p{1.3cm}p{1.3cm}|p{2.8cm}|}
    \hline
    \multicolumn{8}{|c|}{Model \textit{A} vs. Model B}\\
    \hline
    \hline
    \multicolumn{1}{|p{1.1cm}|}{$\Delta_A$-$\Delta_B$} & \multicolumn{1}{|p{1.2cm}}{\centering $>0$} & \centering $\approx0$ & \centering $>0$ & \multicolumn{1}{|p{1.2cm}}{\centering $<0$} & \centering $\approx0$ & \centering $<0$ & \multicolumn{1}{|p{2.8cm}|}{\centering $\approx0$}\\
    \hline
    \multicolumn{1}{|p{1.1cm}|}{$\sigma^2_A-\sigma^2_B$} & \multicolumn{1}{|p{1.2cm}}{\centering $<0$} & \centering $<0$ & \centering $\approx0$ & \multicolumn{1}{|p{1.2cm}}{\centering $>0$} & \centering $>0$ & \centering $\approx0$ & \multicolumn{1}{|p{2.8cm}|}{\centering $\approx0$}\\
    \hline
    \multicolumn{8}{|c|}{Example cases of PWLNs}\\
    \hline
    \textit{A} & \multicolumn{3}{|p{4cm}|}{Single Layer ReLU network with $n_A$ neurons} & \multicolumn{3}{|p{4cm}|}{Deep ReLU network with $L$ layers with $n_l$ neurons in the $l^{th}$ layer ~~~$\forall l = 1, \ldots, L$ and total $n_A$ neurons} & \multicolumn{1}{|p{2.8cm}|}{Single Layer ReLU network with $n_A$ neurons}\\
    \hline
    \textit{B} & \multicolumn{3}{|p{4cm}|}{Deep ReLU network with $L$ layers with $n_l$ neurons in the $l^{th}$ layer ~~~$\forall l = 1, \ldots, L$ and total $n_B$ neurons} & \multicolumn{3}{|p{4cm}|}{Single Layer ReLU network with $n_B$ neurons} & \multicolumn{1}{|p{2.8cm}|}{Single Layer ReLU network with $n_B$ neurons}\\
    \hline
    $\ab{I}{A \leftarrow B}$ & \multicolumn{3}{|p{4cm}|}{$\Big(\frac{1}{c}\Big)^{n.(n_B^{n_0} - n_A^{n_0})}$} & \multicolumn{3}{|p{4cm}|}{$min (\Big(\frac{1}{c}\Big)^{n.(n_B^{n_0} - n_A^{n_0})}, 1)$} & \shortstack{$min (1, $ \\ $\Big(\frac{1}{c}\Big)^{n.(n_B^{n_0} - n_A^{n_0})})$ }\\
    \hline
    $\ub{I}{A \leftarrow B}$ & \multicolumn{3}{|p{4cm}|}{\centering $min\Big(1, $ \newline $\frac{pow\Big(c, ~~{\sum_{s = 0}^{n_0}} {n_A \choose s}\Big)}{\left( \splitfrac{(c^{L}(c-1)^{(n_L - L)n_0})*}{({\displaystyle \prod_{l = 1}^{L-1}}(c - 1)^{(\lfloor\frac{n_l}{n_0}\rfloor n_0)})} \right)}\Big)$} & \multicolumn{3}{|p{4cm}|}{\centering $min\Big(1, $ \newline $ \frac{pow\Big(c, ~~\Big({\displaystyle \sum_{J}} \prod_{l = 1}^{L} {n_l \choose j_l}\Big)\Big)}{c}\Big)$ \newline \newline \centering $\approx 1$} & $min\Big(1, $ \newline $\frac{pow\Big(c, ~\Big({\displaystyle \sum_{s = 0}^{n_0}} {n_A \choose s}\Big)\Big)}{c}\Big)$\\
    \hline
    $\lb{I}{A \leftarrow B}$ & \multicolumn{3}{|p{4cm}|}{\centering $min\Big(1,$ \newline $ \frac{\Large{c}}{pow\Big(c, ~~\Big({\displaystyle \sum_{J}} {\prod_{l = 1}^{L}} {n_l \choose j_l}\Big)\Big)}\Big)$} & \multicolumn{3}{|p{4cm}|}{\centering $min\Big(1,$ \newline \newline $\frac{\left( \splitfrac{(c^{L}(c-1)^{(n_L - L)n_0})*}{({\displaystyle \prod_{l = 1}^{L-1}}(c - 1)^{(\lfloor\frac{n_l}{n_0}\rfloor n_0)})} \right)}{pow\Big(c, ~~{\sum_{s = 0}^{n_0}} {n_B \choose s}\Big)}\Big)$} & \multicolumn{1}{|m{2.8cm}|}{ $\frac{c}{pow\Big(c, ~\Big({\displaystyle \sum_{s = 0}^{n_0}} {n_B \choose s}\Big)\Big)}$}\\
    \hline
\end{tabular}}
\caption{Relative Bias-Variance trade-Off between model \textit{A} and model \textit{B} along with the interpretability bounds for PWLNs for \textbf{complete interpretation}. The two remaining cases represented by (a) $\Delta_A > \Delta_B$ \& $\sigma^2_A > \sigma^2_B$, (b) $\Delta_A < \Delta_B$ \& $\sigma^2_A < \sigma^2_B$ are not possible (In the table, $pow(a, b)$ represents $a^b$)}
\label{table:3}
\end{table*}
{\subsection{Interpretation based on Bias-Variance trade-off}\label{sect:complSpect}
  We consider different cases for models \textit{A} and \textit{B} based on the relative bias-variance trade-off.
Table \ref{table:3} provides a description of the 9 possible cases for \textbf{complete interpretation}, with examples of PWLNs and their bounds.}

This is in contrast to the typical interpretation-by-distillation situation considered in most previous works on interpretability and distillation, where \textit{A} is simpler than \textit{B} i.e. $\Delta_A > \Delta_B$  $\&$ 
{$\sigma^2_A < \sigma^2_B$, where $\Delta$ and $\sigma^2$ represent the Bias and Variance respectively}. 
{The distillation technique \cite{44873} presents a similar idea of transferring knowledge from a complex to a simple network, in contrast to our work which considers the entire spectrum of complexity.
In this typical distillation case (represented in $2^{nd}$ column in Table \ref{table:3}), \textit{A} can be considered as the minimum description length encoding of \textit{B}.} 

\section{Theoretical Bounds on Interpretability for PWLNs}
We determine the theoretical bounds on interpretability with both models \textit{A} and \textit{B} as PWLNs, for \textit{complete interpretation}. We denote the upper, lower and average bounds on the interpretability of model \textit{B} by model \textit{A} as $\ub{I}{A \leftarrow B}$, $\lb{I}{A \leftarrow B}$ and $\ab{I}{A \leftarrow B}$ respectively. Table \ref{table:3} presents the obtained bounds on interpretability for PWLNs. The derivations are demonstrated in the Supplementary Material.
\section{Calculation of Empirical Interpretability}
{Empirical interpretation is determined based on the global surrogate model, as a generalization of distillation \cite{44873}.} 
$H^{emp}_{A \leftarrow B}$ is determined by solving the optimization problem of interpreting model \textit{B} on the set $\hat{\mathcal{D}} = \{(x_i, \hat{y}_i)\}_{i=1}^{n}$, where $\hat{y}_{i}$ represents the prediction of model \textit{B} on the $i^{th}$ example. 
Before interpretation, let model \textit{A} output probability vectors on $\mathcal{D}$ as $P^{in} = \{p^{in}_{1}, p^{in}_{2}, \ldots, p^{in}_{n}\}$, where $p^{in}_{1} \in [0,1]^{c}$.
After interpretation, let model \textit{A} output probability vectors on $\mathcal{D}$ as $P^{fin} = \{p^{fin}_{1}, p^{fin}_{2}, \ldots, p^{fin}_{n}\}$, where $p^{fin}_{1} \in [0,1]^{c}$. 
Here, $in$ and $fin$ represent the initial state (before interpretration) and final state (after interpretation) of model \textit{A} respectively. 
{The initial state of model \textit{A} comes from the initial parameter set assigned to model \textit{A}.}
Figure \ref{fig:empirical} represents the basic idea of the 5-step formulation used for calculating empirical interpretability, using the following color coding:
\begin{figure}[t]
    \centering
    \includegraphics[width =\columnwidth, height=3cm]{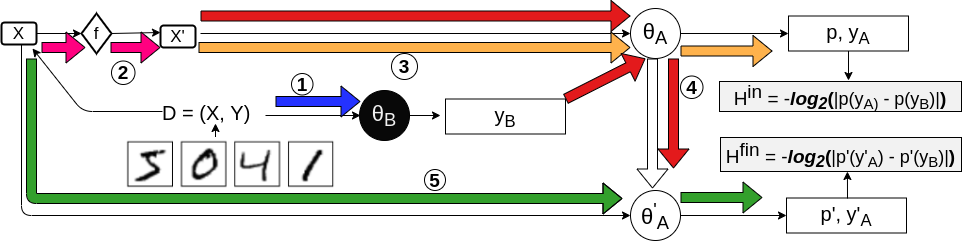}
    \caption{Empirical Interpretation (see text for details on color coding of steps 1-5)}
    \label{fig:empirical}
\end{figure}
\begin{enumerate}
    \item \textcolor{ylgnbu8c7}{Obtain the predictions of black-box model \textit{B} on input $X$.} 
    \item \textcolor{5thcolor}{Transform $X$ to $X'$ as: $X'=\text{crop}(X)$ for section \ref{sect:diffModelInputs} and $X'=X$ otherwise.}
    \item \textcolor{ylorrd3c2}{Obtain predictions and output of model \textit{A} on input $X$ before interpretation.}
    \item \textcolor{paired10c6}{Train model \textit{A} on the predictions of model \textit{B}.}
    \item \textcolor{paired10c4}{Obtain predictions and output of model \textit{A} on input $X$ after interpretation.}
\end{enumerate}
Let the set of predicted classes by model \textit{A} on $\mathcal{D}$ before interpretation be denoted as $\hat{Y}^{in}_{A} = \{\hat{y}^{in}_{1}, \hat{y}^{in}_{2}, \ldots, \hat{y}^{in}_{n}\}$, where $\hat{y}_{i}^{in} = argmax_{j \in \{1, \ldots, c\}}(p^{in}_{ij})$. 
Similarly, the set of predicted classes by model \textit{A} on $\mathcal{D}$ after interpretation is denoted as $\hat{Y}_{A}^{fin} = \{\hat{y}^{fin}_{1}, \hat{y}^{fin}_{2}, \ldots, \hat{y}^{fin}_{n}\}$, where $\hat{y}^{fin}_{i} = argmax_{j \in \{1, \ldots, c\}}(p^{fin}_{ij})$.

Consider the $i^{th}$ sample. Based on the above formulation; $p^{in}_{i,\hat{y}_{i}}$ represents the probability assigned by model \textit{A} to the predicted class of model \textit{B} (i.e. $\hat{y}_{i}$), before the process of interpretation. 
Similarly, $p^{fin}_{i,\hat{y}_{i}}$ represents the corresponding probability after the process of interpretation. 
On the other hand, $p^{in}_{i,\hat{y}^{in}_{i}}$ and $p^{fin}_{i,\hat{y}^{fin}_{i}}$ represent the maximum probability values assigned to any class by \textit{A} before and after the process of interpretation respectively. 

Now, we define the entropies $H^{emp}_{A \not\leftarrow B}$ and $H^{emp}_{A \leftarrow B}$. 
The general notion of entropy considers $p^{in}_{=} = P(p^{in}_{i,\hat{y}_{i}} = p^{in}_{i,\hat{y}^{in}_{i}})$ and $p^{in}_{\neq} = P(p^{in}_{i,\hat{y}_{i}} \neq p^{in}_{i,\hat{y}^{in}_{i}})$. 
$p^{fin}_{=}$ and $p^{fin}_{\neq}$ can be defined in a similar way. 
Then, the entropies are defined as:
\begin{equation}
\begin{split}
    H^{emp}_{A \not\leftarrow B} &= -(p^{in}_{=})\log_2(p^{in}_{=}) - (p^{in}_{\neq})\log_2(p^{in}_{\neq})\\
    H^{emp}_{A \leftarrow B} &= -(p^{fin}_{=})\log_2(p^{fin}_{=}) - (p^{fin}_{\neq})\log_2(p^{fin}_{\neq})
\end{split}
\end{equation}
We define the above entropies in a slightly different way. We calculate the entropies using differences in the probabilities calculated earlier (as a lower difference implies more information gain), as:
\begin{equation}  \label{eq:procentropies}
\begin{split}
    H^{emp}_{A \not\leftarrow B} &= \sum\nolimits_{i = 1}^{n}-\log_2(|p^{in}_{i,\hat{y}_{i}} - p^{in}_{i,\hat{y}^{in}_{i}}|)\\
    H^{emp}_{A \leftarrow B} &= \sum\nolimits_{i=1}^{n} -\log_2(|p^{fin}_{i,\hat{y}_{i}} - p^{fin}_{i,\hat{y}^{fin}_{i}}|)
\end{split}
\end{equation}

The empirical interpretability can now be formulated based on the definition of interpretability in Equation (\ref{interpretability}) and entropies defined as in Equation (\ref{eq:procentropies}).

\section{Experimental Setup and Results}
Through these experiments, we demonstrate the applicability of our proposed interpretation-by-distillation framework in diferent supervised classification scenarios. 
Our experiments demonstrate the stability of our interpretation-by-distillation framework as well as its conformity to human understanding, while interpreting state-of-the-art deep learning models.
\subsection{Overview of Experimental Setup}
\begin{figure*}[h]
    \centering
    \includegraphics[width=0.9\textwidth]{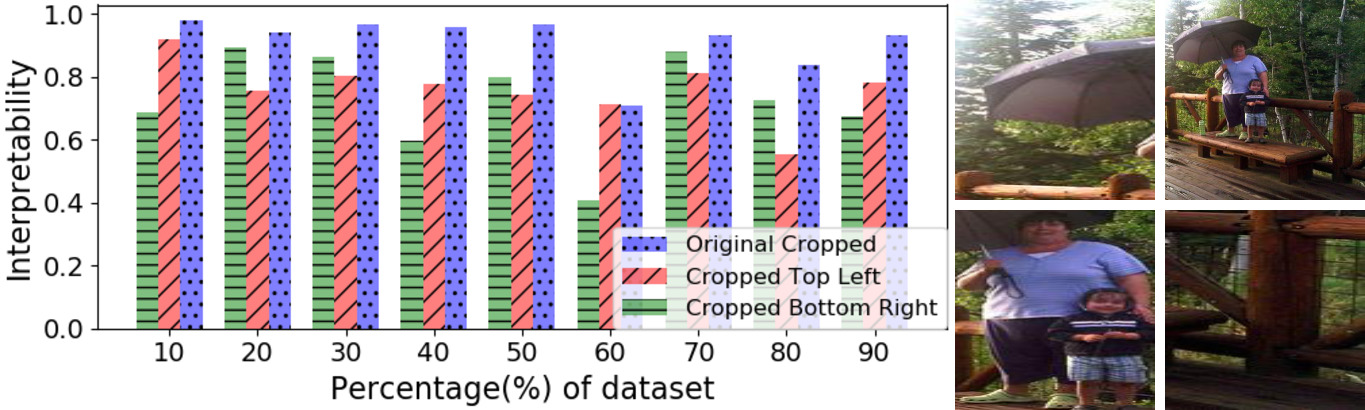}
    \caption{\textbf{Left}: Empirical Interpretability when an InceptionV3 network trained on the Stanford40 dataset is interpreted by another InceptionV3 network trained on different cropped versions of the same set of images. \textbf{Right}: An example from the Stanford40 dataset labelled as "Holding An Umbrella", showing the original image (top right), Original Cropped Image (bottom left), Cropped Top Left Image (top left) and Cropped Bottom Right Image (bottom right).}
    \label{fig:cropped_interpret}
\end{figure*}
In Experiment \ref{sect:diffModelInputs}, our quantification conforms with the human understandable visual explanations of the predictions of the state-of-the-art InceptionV3 \cite{7780677} network .
In Experiment \ref{sect:explaining_black_box} we break down the complex structure of a MiniVGGNet in the form of a human understandable ensemble of simpler models, through our interpretation-by-distillation framework.
In Experiment \ref{sect:number_samples}, we demonstrate the stability of our interpretation formulation for PWLNs (based on our theoretical formulation). 
\textbf{Dataset:} In Experiment \ref{sect:diffModelInputs}, we use the Stanford40 dataset \cite{6126386}, which contains 9072 RGB images corresponding to 40 different human actions.
We use the popular MNIST \cite{lecun-mnisthandwrittendigit-2010}
and Fashion-MNIST \cite{xiao2017/online} datasets for all other experiments. Both datasets contain 60,000 and 10,000 grayscale images (of size 28$\times$28) as part of the train and test sets respectively, with 10 output classes.\\
\textbf{Data Pre-Processing:} We normalize the data using a min-max normalization and perform a 5-fold cross validation split on the training set, considering 4 folds for interpretation and the remaining fold for cross-validation. Empirical interpretability is calculated on the test set. \\
\textbf{Notations:} In the experiments below, $\gamma$ represents the batch size and $\alpha$ represents the learning rate. Also, the term "Piece-Wise Linear Neural Networks with ReLU activations" is abbreviated as PWLN-R.\\
We implement the experimental setup in Python using the \textit{Tensorflow} and \textit{Keras} libraries. {The experiments were conducted on a \textit{NVIDIA Tesla V100} GPU node with 192GB RAM and CentOS Linux. 
\footnote{The code will be made publicly available upon acceptance.}}
{\subsection{Explaining the predictions}\label{sect:diffModelInputs}
Visual explanation of the predictions of black-box models is key to human understanding of these models for various tasks. 
While our interpretation-by-distillation framework is completely theoretical; through this experiment, we provide a connection between our interpretability measure and human understanding of the decision structure of black-box models. We determine the parts of the input image which affect the classifier prediction in the maximum capacity, thus explaining where the classifier "looks" in the image for the task-at-hand.\\
\textbf{Model Configurations:} Both models \textit{A} and \textit{B} are InceptionV3 networks, pre-trained on ImageNet. We add a global-average pooling and two dense layers at the end for fine-tuning. \textit{B} is trained on the original Stanford40 images while \textit{A} is trained on cropped portions of the images in the Stanford40 dataset. We use the annotations and bounding boxes for the associated objects provided with the Stanford-40 dataset.
This experiment falls under the relative complexity case of both models having similar complexity in Table \ref{table:3}.\\
\textbf{Hyperparameters and Pre-Processing:} $\gamma = 256$, $\alpha=0.01$, $no\_of\_epochs=150$ for both models. For 5 epochs, the models are trained using \textit{RMSProp} optimizer, and then \textit{SGD} optimizer is used for 150 epochs. 
We also augment the dataset using Keras libraries.
The first 249 layers of the models are kept fixed during training.
All cropped images are resized to a 200$\times$200 shape across all the 3 channels.\\
\textbf{Explanation:} Figure \ref{fig:cropped_interpret} demonstrates the behaviour of interpretability when the Stanford40 images are cropped in 3 different fashions. A higher interpretability is obtained when the cropped images contain the objects in focus (e.g.- bottom left image in Figure \ref{fig:cropped_interpret} Right, containing the objects "Person" and "Umbrella" for the action "Holding An Umbrella"). The interpretability is lower when cropping is done from the top left and bottom right corners (with the same size as that of the bounding box) (e.g. - see Fig. \ref{fig:cropped_interpret} Right) as shown by the red and green bars respectively in Figure \ref{fig:cropped_interpret}. Note that, the accuracy remains the same for all the three different cropping cases, at around 70\%.}
\begin{figure}[h]
    \centering
    \includegraphics[width=0.9\columnwidth, keepaspectratio]{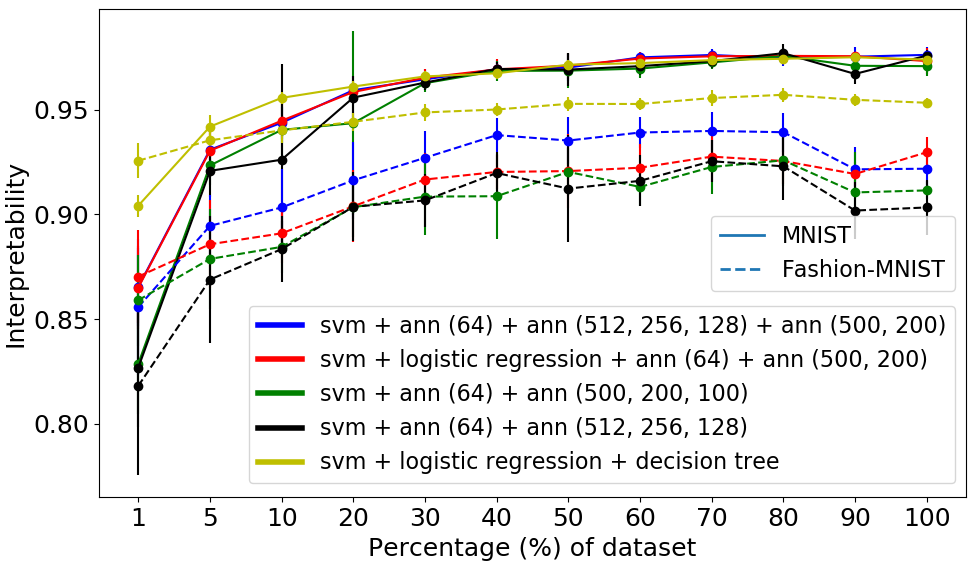}
    \caption{Evaluation of Empirical Interpretability on MNIST and Fashion MNIST. Different ensembles (\textit{A}) are used to interpret a MiniVGGNet (\textit{B}).}
    \label{fig:SamplesVsInterp_ensemble}
\end{figure}
\begin{figure*}[ht]
    \centering
    \includegraphics[width=0.9\textwidth, height=4cm]{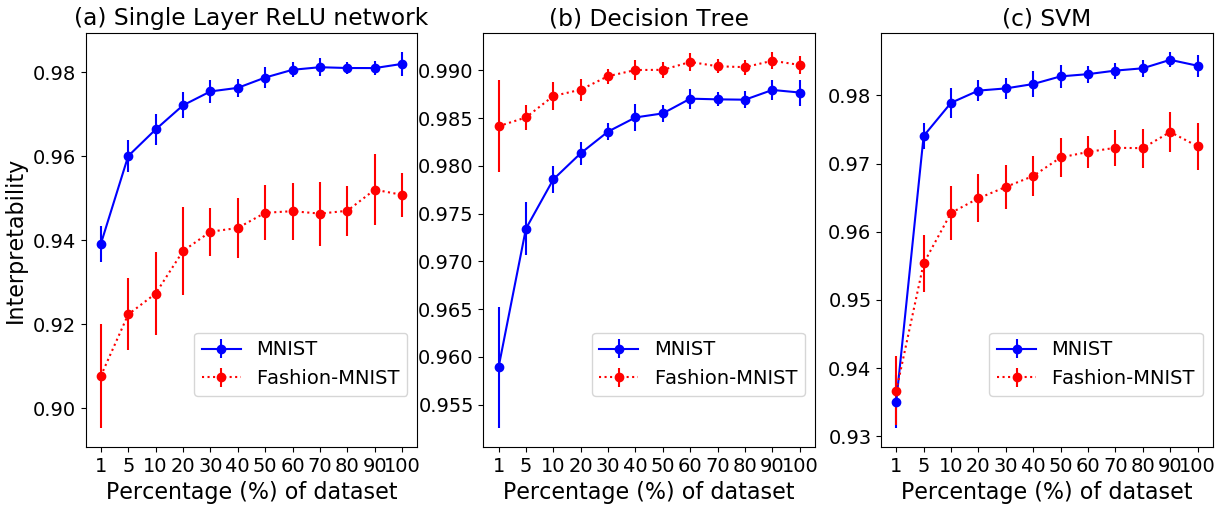}
    \caption{Effect of the number of samples on Empirical Interpretability, when a 1-layer PWLN-R(a), Decision Tree(b) and an SVM(c) are used to interpret a 4-layer PWLN-R.}
    \label{fig:SamplesVsInterp}
\end{figure*}

{\subsection{Explaining the Black Box Model}\label{sect:explaining_black_box}
Using ensembles of human understandable models to interpret black-box models encourages the breakdown of the complex decision mechanism of the black-box model into a human understandable form.
Further, using an ensemble to interpret black-box model \textit{B} removes the bias of the choice of model \textit{A} prevalent in previous model-based interpretation mechanisms.\\
\textbf{Model Configurations:} We take \textit{B} as a MiniVGGNet, and use different ensembles as \textit{A}, to interpret \textit{B}. 
This experiment falls under the relative complexity case where \textit{A} is simpler than \textit{B} in Table \ref{table:3}.\\
\textbf{Hyperparameters:} $\gamma=128$, $\alpha = 1e-3$. Both models use \textit{Adam} optimizer and are trained for 150 epochs each\\
\textbf{Explanation:} Figure \ref{fig:SamplesVsInterp_ensemble}, demonstrates the interpretability of a CNN in terms of a simple ensemble, particularly the ensemble of SVM, Logistic Regression and Decision Tree, which interprets better than other more complex ensembles on both datasets.}

\subsection{Effect of the number of Queries}\label{sect:number_samples}
As explained in Section \ref{sect:formulation}, the process of \emph{complete interpretation} is relaxed into the process of \emph{empirical interpretation} for practical purposes.
Due to finite queries in \emph{empirical interpretation}, the choice of the dataset affects empirical interpretability.
However, as the number of queries is increased, the \emph{empirical interpretation} mechanism moves towards \emph{complete interpretation}.
As \emph{complete interpretation} is achieved, the bias introduced by the choice of the dataset is removed, thus suggesting the convergence of interpretability values with increasing number of queries.
Hence, we demonstrate the effect of the number of queries on the empirical interpretability (while keeping the size of test set for interpretability calculations fixed) and demonstrate convergence of interpretability with increasing number of queries.\\
\textbf{Model Configurations:} We fix model \textit{B} as a 4-layer PWLN-R with 512, 256, 128 and 64 neurons in its 4 layers respectively. 
We use three different architectures for model \textit{A}: a 1-layer PWLN-R with 256 hidden neurons, a Decision Tree with \textit{Gini} criterion and an SVM with \textit{RBF} kernel.\\
\textbf{Hyperparameters:} $\gamma = \mathit{128}$, $\alpha = \mathit{1e-3}$. 
We train both \textit{B} and \textit{A} for 30 epochs each, using \textit{Adam} optimizer and \textit{Truncated Normal} initializer 
We determine the mean and deviation over three runs for each point.\\
\textbf{Explanation:} Figure \ref{fig:SamplesVsInterp} demonstrates that by increasing the number of queries (1-100\% of the dataset) for \textit{A} to perform \textit{empirical interpretation} on \textit{B}, it is better able to match the decision boundary of the black-box model \textit{B} and a stable value of interpretability is achieved (demonstrated by the decreasing deviations of interpretability values).
As the number of query points increases to 100\% of the data, we observe that the interpretability converges.

All the models perform better (in terms of fidelity) on MNIST as compared to Fashion-MNIST, owing to higher complexity of Fashion-MNIST. However as in Fig. \ref{fig:SamplesVsInterp}(b), despite obtaining higher accuracy on MNIST, the interpretability is lower for Decision Tree classifier, thus showing that our interpretability formulation is decoupled from accuracy. 

\section{Discussion and Future Work}
Our interpretation-by-distillation framework provides a novel interpretability definition from an information-theoretic perspective, with the aim of providing policymakers an unbiased interpretability estimate.
We also provide the first theoretical bounds on the interpretability of ReLU networks and demonstrate stability and conformity of our proposed formulation to human understanding.

The current theoretical interpretability estimates are quite far from the obtained empirical estimates (due to the exponential nature of the complexity estimates from previous literature), but present a good starting point for developing tighter estimates. As tighter complexity bounds will be derived in future, the theoretical measures will closely follow the empirical ones.

In our future work, we also plan to explore the effect of complexity of the dataset used for interpretation. 
As the dataset complexity increases, it becomes increasingly difficult to emulate the decision boundary of the black box model. 
We plan to take this into consideration for our interpretability formulation.
Further, our model entropy definition currently only considers the arrangement of the decision boundary. Our future work would focus on incorporating the shape of the decision boundary as well into our entropy definition. 

Our future work would aim to derive tight upper and average bounds on the entropies of PWLNs, just like the tight lower bounds derived in the current work.
Further, pur proposed theoretical interpretability bounds are limited to PWLNs, since previous literature has explored complexity bounds for only PWLNs. 
As and when the complexities for networks like CNNs are defined, our work can be extended to these networks as well.

The current work presents a first direction towards a new definition for interpretability, dissociated from human understanding. This presents researchers with a previously unexplored notion of quantifying interpretability using information theory which we hope inspires other researchers to explore further.

\bibliography{main.bib}

\newpage
\newpage
\section{Supplementary Material}
\beginsupplement
\subsection{Complexity of PWLNs}
Consider two PWLNs with ReLU activations, say $P$ and $Q$. 
$P$ is a \emph{single-layer} PWLN with $h$ hidden neurons. $Q$ is a \emph{deep} PWLN with $L$ layers and $n_l$ neurons in its $l^{th}$ hidden layer. Let the total number of hidden neurons in the models $P$ and $Q$ be $n_P$ and $n_Q$ respectively. Then, the complexity bounds are summarized as: 
\begin{itemize}
    \item For model $P$, $\ub{C}{P}$ is given by $\sum_{s = 0}^{n_0}$ ${h} \choose{s}$ \cite{zaslavsky1975}, while $\lb{C}{P} = 1$.
    \item For model $Q$, $\ub{C}{Q}$ is given by $\sum_{J} \prod_{l = 1}^{L}$ $n_{l} \choose{j_{l}}$, where $J = \{(j_1, j_2, \ldots, j_{L}) \in \mathbb{Z}^{L} : 0 \leq j_l \leq min(n_0, n_1 - j_1, \ldots, n_{l-1} - j_{l-1}, n_l) \forall l = 1, \ldots, L\}$\cite{serra2018bounding}. $\mathbb{Z}$ denotes the set of integers.
     \item For model $Q$, $\lb{C}{Q}$ is given by $\Big(\prod_{i = 1}^{L-1} \lfloor\frac{n_i}{n_0}\rfloor^{n_0} \Big)\sum_{j = 0}^{n_0}$ ${n_L} \choose{j}$ \cite{NIPS2014_5422}. We do not consider the lower bound given by \cite{serra2018bounding} as it is more restrictive, requiring $n_l \geq 3n_0$ as compared to $n_l \geq n_0$ in \cite{NIPS2014_5422}, $\forall l = \{1, 2, \ldots, L\}.$
    \item $\ab{C}{P}$ and $\ab{C}{Q}$ are given by $(n_P)^{n_0}$ and $(n_Q)^{n_0}$ respectively, when $n_0 > 1$. For $n_0 = 1$, $\ab{C}{P}$ and $\ab{C}{Q}$ are given by $n.T.n_P$ and $n.T.n_Q$ respectively, where $T$ represents the number of breakpoints in the non-linearity of the activation function of the neural network (for ReLU, $T = 1$). \cite{complexityAverage}.
\end{itemize}
\subsection{Interpretability Formulation}
As defined in the interpretation mechanism, we have two models $A$ and $B$ where $B$ is the black-box model which is being interpreted by the known model $A$. Then, we define the entropy $H_{A \not\leftarrow B}$ as the number of mappings in the input space identified by model $B$ but not identified by model $A$, before the process of \emph{complete interpretation}. Similarly, we define $H_{A \leftarrow B}$ as the number of mappings in the input space identified by model $B$ but not identified by model $A$, after the process of \emph{complete interpretation}. Hence,
\begin{align*}
    H_{A \not\leftarrow B} &= H_B\\
    H_{A \leftarrow B} &= \max(0, H_B - H_A)\\
    I_{A \leftarrow B} &= \frac{H_B - \max(0, H_B - H_A)}{H_B}
\end{align*}

\subsection{Model Entropy}
The derivation here is based on a similar concept as used by Montufar et al. \cite{NIPS2014_5422} for the derivation of the lower bound on the maximal number of linear regions for deep rectifier networks. The derivation of \cite{NIPS2014_5422} clearly demonstrates that every linear region identified in the input space by the PWLN maps to the same region in the output space, hence for PWLNs our derivation holds as for PWLNs, cells are the same as linear regions. We demonstrate the derivation for rectifier networks, but the result can be extended to other PWLNs as well using a similar idea.

Consider a L-layered deep neural network composed of ReLU activations and containing $n_l$ neurons in its $l$-th layer. Let $n_0$ represent the number of input variables, where $n_l > n_0, \forall l = \{1, 2, \ldots, L\}$. Now, partition the set of $n_l$ neurons in the $l^{th}$ layer into $n_0$ subsets, each with cardinality $p = \lfloor\frac{n_l}{n_0}\rfloor$. For simplification, we assume $n_0$ divides $n_l$ and there are no remaining neurons, however, the construction can be easily modified for the case of remaining neurons as well.

As demonstrated in \cite{NIPS2014_5422}, an alternating sum of rectifier units divides the input space into equal-length segments. If we consider the rectifier units in the $j^{th}$ subset, we can choose the input weights and biases of the units in this subset such that they are only sensitive to the $j^{th}$ coordinate of the input $x$ and the output activations of these units are given in \cite{NIPS2014_5422}.

\begin{figure*}[t]
    \centering
    \includegraphics[width=0.7\textwidth]{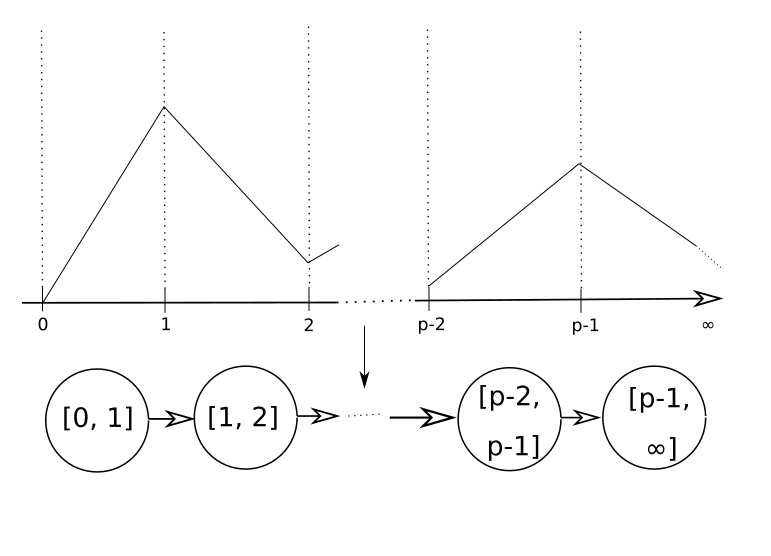}
    \caption{Linear Regions in each input dimension represented as a path graph}
    \label{fig:Linear Chain}
\end{figure*}
The alternating sum of these rectifier units of the $j^{th}$ subset produces a new function $\Tilde{h}_{j}(x) = h_1(x) - h_2(x) + \ldots + (-1)^{p-1}h_p(x)$ which effectively acts on only the scalar input $x_j$.

This construction identifies p cells in each coordinate of the input $x$, given as
$[0, 1]$ $, [1, 2], \ldots, [p-1, \infty)$. Now, considering all the $n_0$ subsets each of which operates on distinct dimensions of the input, we get a total of $p^{n_0}$ hypercubes which all map into the same output space. 

As illustrated in the construction in \cite{NIPS2014_5422}, the identified input cells are continuous along each dimension. Thus, the cells can be identified as forming a graph, where each cell represents a vertex and any two continuous/adjacent cells are connected directly via an edge. The $p$ linear cells thus form a linear chain along each dimension as shown in Figure \ref{fig:Linear Chain}, which when modelled graphically represents a path graph.\\
Based on the definition of the entropy of a supervised classification model as given in Definition \ref{entropy}, we can visualize the problem of assigning classes to the cells in the input space as a \textbf{\emph{graph coloring problem}}, where there are $p$ vertices and $c$ colors (representing the $c$ classes). For the graph coloring problem, the chromatic polynomial \cite{10.2307/1967597} defines a graph polynomial which counts the number of colorings possible for the graph as a function of the number of colors.

Having modelled the cells in each dimension as a path graph, our problem is essentially converted into a path-graph-coloring problem with $p$ vertices and $c$ colors. For a path graph, the chromatic polynomial is given by $P(G, k) = k(k - 1)^{(n - 1)}$, where $n$ is the number of vertices. So, the number of possible colorings for the defined path graph is given as $c(c - 1)^{(p - 1)}$, for each subset defined in the $l$-th layer. Extending this for all the $n_0$ subsets, we have the total number of possible colorings as $c(c - 1)^{(p-1)n_0}$. This is true for layers $l = \{1, 2, \ldots, L - 1\}$. Hence, the total number of possible colorings on the cells in the input space, upto layer $L - 1$ is given by ${\displaystyle \prod_{l = 1}^{L-1}}c(c - 1)^{(\lfloor\frac{n_l}{n_0}\rfloor - 1)n_0}$.

For the last layer $l = L$, the number of linear cells formed is given by $O(n_L^{n_0})$. 
Hence, using a similar construction as earlier, we can say that the last layer defines a total of $n_L$ linear cells (forming a path graph), in each of the $n_0$ input dimensions, so the total number of possible colorings induced by the last layer is given by $c(c-1)^{(n_L - 1)n_0}$. As a result, the total number of possible colorings in the input space formed by the entire network is given by $({\displaystyle \prod_{l = 1}^{L-1}}c(c - 1)^{(\lfloor\frac{n_l}{n_0}\rfloor - 1)n_0})*(c(c-1)^{(n_L - 1)n_0})$.

\subsection{Interpretability of PWLNs based on the relative Bias-Variance Trade-Off}
Let model $A$ be a 1-layer PWLN with ReLU activations, with $h$ hidden neurons, and model $B$ be a deep PWLN with ReLU activations, with $L$ layers and $n_l$ neurons in its $l^{th}$ hidden layer. Both models take an $n_0$-dimensional input. Let the total number of neurons in the models $A$ and $B$ be given by $n_A$ and $n_B$ respectively. Based on the complexity bounds defined in Section \ref{sect:background}, we can obtain the values given in Table \ref{tabel:Comp&Entropy}.\\
\begin{table}[h]
    \centering
    \caption{Bounds on Complexity and Entropy for PWLNs with ReLU activations. (In the table, $pow(a, b)$ represents $a^b$ and $J = \{(j_1, j_2, \ldots, j_{L}) \in \mathbb{Z}^{L} : 0 \leq j_l \leq min\{n_0, n_1 - j_1, \ldots, n_{l-1} - j_{l-1}, n_l\} \forall l = 1, \ldots, L\}$) }
    \def\arraystretch{1.5}
    \begin{tabular}{|c|>{\centering\arraybackslash}p{2.5cm}|>{\centering\arraybackslash} p{4cm}|}
        \hline
        \bf & \bf Model $P$ as $A$ & \bf Model $P$ as $B$ \\
        \hline
        $\ub{C}{P}$ & \centering ${\sum_{s = 0}^{n_0}}$ ${h} \choose{s}$ & $\sum_{J} \prod_{l = 1}^{L}$ $n_{l} \choose{j_{l}}$\\
        \hline
        $\lb{C}{P}$ & 1 & $\Big({\prod_{i = 1}^{L-1}} \lfloor\frac{n_i}{n_0}\rfloor^{n_0} \Big){\sum_{j = 0}^{n_0}}$ ${n_L} \choose{j}$ \\
        \hline
        $\ab{C}{P}$ & $(n_A)^{n_0}$ & $(n_B)^{n_0}$ \\
        \hline
        $\ub{H}{P}$ & $pow\Big(c,{\sum_{s = 0}^{n_0} {h \choose s}}\Big)$ &  $pow\Big(c,\sum_{J} \prod_{l = 1}^{L} {n_{l} \choose j_{l}}\Big)$\\
        \hline
        $\lb{H}{P}$ & $(c(c-1)^{(h - 1)n_0})$ & $({\prod_{l = 1}^{L-1}}(c - 1)^{(\lfloor\frac{n_l}{n_0}\rfloor n_0)})$ \newline $*(c^L(c-1)^{(n_L - L)n_0})$ \\
        \hline
        $\ab{H}{P}$ & $pow\Big(c,(n_A)^{n_0}\Big)$ & $pow\Big(c,(n_B)^{n_0}\Big)$ \\
        \hline
    \end{tabular}
    \label{tabel:Comp&Entropy}
\end{table}
Thus, the various bounds on interpretability, $\ub{I}{A \leftarrow B}$, $\lb{I}{A \leftarrow B}$ and $\ab{I}{A \leftarrow B}$ can be defined as:
\begin{align*}
    \ub{I}{A \leftarrow B} &= \frac{\lb{H}{B} - \max(0, \lb{H}{B} - \ub{H}{A})}{\lb{H}{B}}\\
     &= \min(1, \frac{\ub{H}{A}}{\lb{H}{B}})
\end{align*}
\begin{align*}     
    \lb{I}{A \leftarrow B} &= \frac{\ub{H}{B} - \max(0, \ub{H}{B} - \lb{H}{A})}{\ub{H}{B}}\\
    &= \min(1, \frac{\lb{H}{A}}{\ub{H}{B}})
\end{align*}
\begin{align*}
    \ab{I}{A \leftarrow B} &= \frac{\ab{H}{B} - \max(0, \ab{H}{B} - \ab{H}{A})}{\ab{H}{B}}\\
    &= \min(1, \frac{\ab{H}{A}}{\ab{H}{B}})
\end{align*}
The formulae in Table \ref{table:3} are simple applications of the formulae presented above.

\subsection{Experiment: Effect of the Optimizer}
\begin{figure}[h]
    \centering
    \includegraphics[width=0.9\columnwidth]{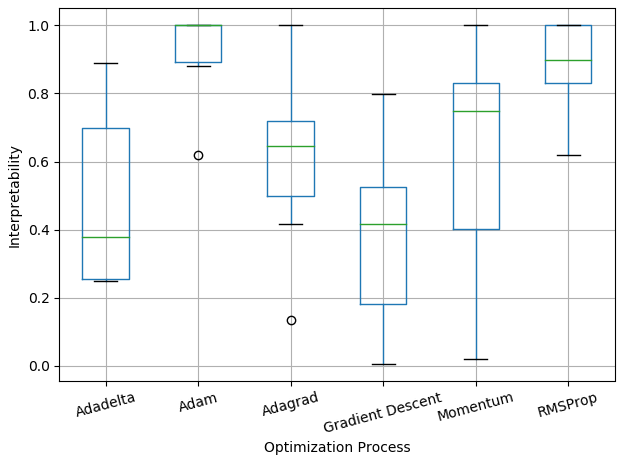}
    \caption{Empirical Interpretability between models \textit{A} and \textit{B} computed by different optimizers.}
    \label{fig:optimProcess}
\end{figure}
We study the effect of different optimization techniques (an implementation choice) for the optimization process (given at the end of Section \ref{sect:formulation}) over our empirical interpretability.\\
\textbf{Model Configurations}: Models \textit{A} and \textit{B} have the same configurations as in Section \ref{sect:number_samples}. 
The experiment is performed on the MNIST dataset.
The box plots are constructed using $\alpha = \{1e-4, 3e-4, 1e-3, 3e-3, 1e-2, 3e-2, 0.1, 0.3, 1\}$.\\
\textbf{Explanation:} The box plots in Figure \ref{fig:optimProcess} demonstrates that both RMSProp and Adam are very stable and perform better than other optimizers, when used by model \textit{A} to interpret model \textit{B}. Further, due to a finite number of queries, different optimizers affect the optimization process differently. 

\subsection{Experiment: Explaining PWLNs}
\begin{figure}[h]
    \centering
    \includegraphics[width=0.9\columnwidth, keepaspectratio]{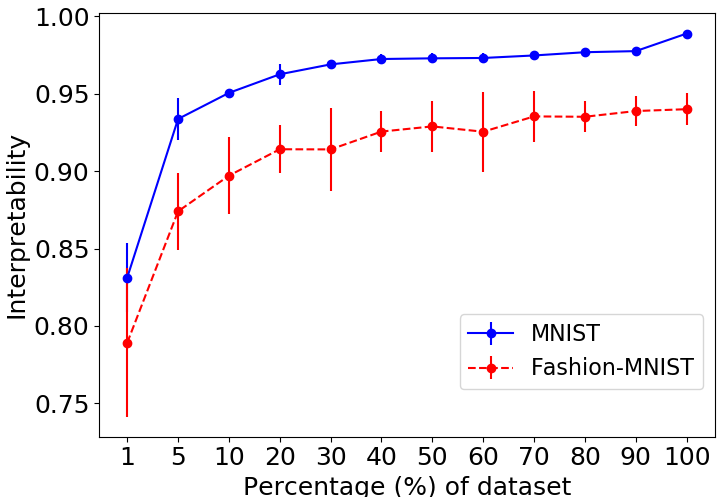}
    \caption{Evaluation of Empirical Interpretability on MNIST and Fashion MNIST. An ensemble of an SVM, Decision Tree and 1-layer PWLN-R (\textit{A}) is used to interpret a 4-layer PWLN-R (\textit{B})}
    \label{fig:ensemble_PWLN}
\end{figure}

We use an ensemble to interpret PWLNs, which breaks down their complex decision structures into ensembles of simpler models.
This is an extension to experiment \ref{sect:explaining_black_box}.\\
\textbf{Model Configurations:} We fix model \textit{B} as a 4 layer PWLN-R with 512, 256, 128 and 64 neurons in its 4 layers respectively. We use an ensemble of an SVM, Decision Tree and a 1-layer PWLN-R with 256 neurons, with model averaging technique, as model \textit{A}.\\
\textbf{Hyperparameters:} $\gamma=128$ and $\alpha = 1e-3$. Both models are trained for 30 epochs using \textit{Adam} optimizer for both models. \\

\textbf{Explanation:} Figure \ref{fig:ensemble_PWLN} shows that the decision structure of the black-box 4 layer PWLN-R can be explained in terms of the simplified decision structure of the ensemble \textit{A} with high interpretability.

{\subsection{Double Descent Behaviour}
\begin{figure}[t]
    \centering
    \includegraphics[width=0.9\columnwidth, keepaspectratio]{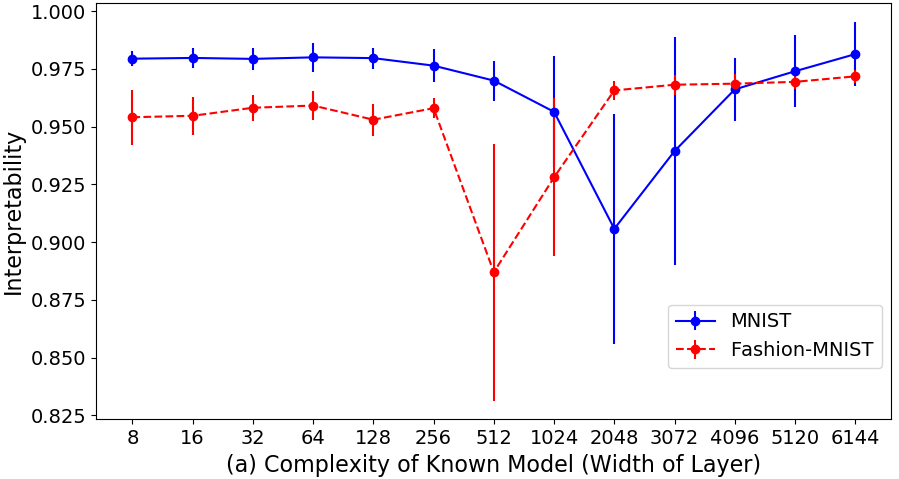}
    \caption{Double Ascent behaviour when a 1-layer PWLN-R with 512 neurons is interpreted by a 1-layer PWLN-R with different number of hidden neurons.}
    \label{fig:combined}
\end{figure}
The behaviour of modern deep learning methods is quite at odds with the classical U-shaped risk curve defined by the bias-variance trade-off. 
Modern deep learning architectures, on the contrary, tend to obtain high accuracy on both the test and train sets. \cite{doubledescent} demonstrates the double descent curve, which better explains the behaviour of modern deep learning networks.
We determine whether our interpretation-by-distillation mechanism is in conformance with the equivalent double ascent curve of interpretability.\\
\textbf{Model Configurations:} \textit{B} is a 1-layer PWLN-R with 512 hidden neurons, \textit{A} as a 1-layer PWLN-R with increasing number of hidden neurons in the order $[8, 16, \ldots, 5120, 6144]$. This covers the entire relative complexity spectrum between \textit{A} and \textit{B}.\\
\textbf{Hyperparameters:} $\gamma=128$, $\alpha=1e-3$. We train both models for 20 epochs with Adam optimizer, \textit{Truncated Normal kernel initializer}, all biases initialized to 1 and no regularizers. We obtain the mean and deviation over three runs.\\
\textbf{Explanation:} Figure \ref{fig:combined} demonstrates the conformity of our interpretability formulation with the double ascent behaviour. This shows that our formulation replicates behaviours that are fundamental in modern deep learning.}

\end{document}